\documentclass[letterpaper]{article} 
\usepackage{aaai2026}  
\usepackage{times}  
\usepackage{helvet}  
\usepackage{courier}  
\usepackage[hyphens]{url}  
\usepackage{graphicx} 
\usepackage{natbib}  
\usepackage{caption} 
\usepackage{algorithm}
\usepackage{algorithmic}

\usepackage{amssymb}
\usepackage{amsmath}
\usepackage{color}
\usepackage{multirow}
\usepackage{enumitem}
\usepackage{float}
\usepackage{booktabs}
\usepackage{subcaption}
\usepackage{graphicx}
\usepackage{epstopdf}
\usepackage{adjustbox}
\usepackage{newfloat}
\usepackage{listings}
\DeclareCaptionStyle{ruled}{labelfont=normalfont,labelsep=colon,strut=off} 
\floatstyle{ruled}
\newfloat{listing}{tb}{lst}{}
\floatname{listing}{Listing}
\title{YOLO-PRO: Enhancing Instance-Specific Object Detection with Full-Channel Global Self-Attention}
\author{
    Lin Huang\textsuperscript{\rm 1}\textsuperscript{,}\textsuperscript{\rm 3}\textsuperscript{,}\textsuperscript{\rm 4}, Yujuan Tan\textsuperscript{\rm 1}\textsuperscript{,}\thanks{Corresponding Author.}, Weisheng Li\textsuperscript{\rm 2}, Shitai Shan\textsuperscript{\rm 3}\textsuperscript{,}\textsuperscript{\rm 4}, Liu Liu\textsuperscript{\rm 5}, Linlin Shen\textsuperscript{\rm 6},Jing Yu\textsuperscript{\rm 1},Yue Niu\textsuperscript{\rm 1}
}
\affiliations{

    \textsuperscript{\rm 1}Chongqing University\\
    \textsuperscript{\rm 2}Chongqing University of Posts and Telecommunications\\
    \textsuperscript{\rm 3}Inspur Yunzhou Industrial Internet Co., Ltd\\
    \textsuperscript{\rm 4}Guoqi Zhimo (Chongqing) Technology Co., Ltd.\\
    \textsuperscript{\rm 5}China Academy of Information and Communications Technology\\
    \textsuperscript{\rm 6}Shenzhen University\\
    h72001346@163.com, tanyujuan@cqu.edu.cn, liws@cqupt.edu.cn, shansht@inspur.com, 244319450@qq.com, llshen@szu.edu.cn, 20201401010@cqu.edu.cn, yuen@cqu.edu.cn
}

\usepackage{bibentry}

\begin{document}

\maketitle

\begin{abstract}

This paper addresses the inherent limitations of conventional bottleneck structures (diminished instance discriminability due to overemphasis on batch statistics) and decoupled heads (computational redundancy) in object detection frameworks by proposing two novel modules: the Instance-Specific Bottleneck with full-channel global self-attention (ISB) and the Instance-Specific Asymmetric Decoupled Head (ISADH). The ISB module innovatively reconstructs feature maps to establish an efficient full-channel global attention mechanism through synergistic fusion of batch-statistical and instance-specific features. Complementing this, the ISADH module pioneers an asymmetric decoupled architecture enabling hierarchical multi-dimensional feature integration via dual-stream batch-instance representation fusion. Extensive experiments on the MS-COCO benchmark demonstrate that the coordinated deployment of ISB and ISADH in the YOLO-PRO framework achieves state-of-the-art performance across all computational scales. Specifically, YOLO-PRO surpasses YOLOv8 by 1.0-1.6\% AP (N/S/M/L/X scales) and outperforms YOLO11 by 0.1-0.5\% AP in critical N/M/L/X groups, while maintaining competitive computational efficiency. This work provides practical insights for developing high-precision detectors deployable on edge devices.

\end{abstract}


\section{Introduction}

\begin{figure}[H]
\centering
    \begin{subfigure}[b]{0.42\linewidth}
        \includegraphics[width=\textwidth]{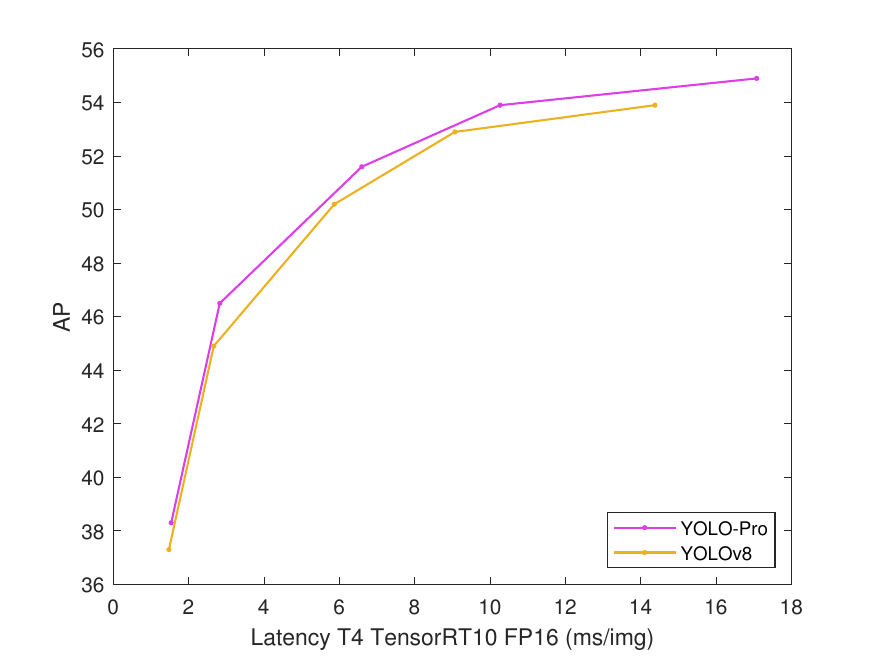}
        \caption{AP and Latency}
    \end{subfigure}
    \hfill
    \begin{subfigure}[b]{0.42\linewidth}
        \includegraphics[width=\textwidth]{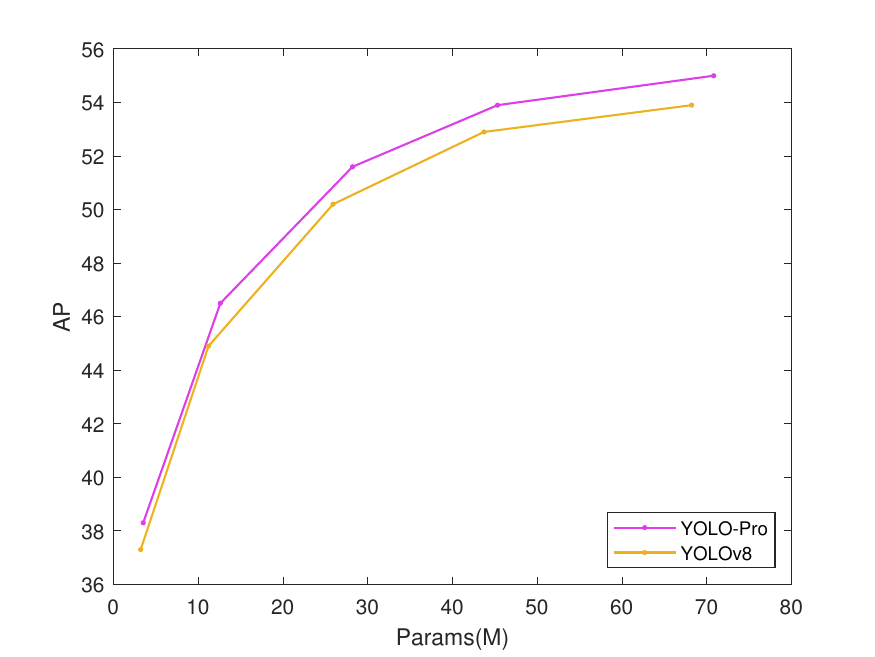}
        \caption{AP and Params}
    \end{subfigure}
    \caption{Comparison of the proposed YOLO-PRO with YOLOv8.}
    \label{fig:total}

\end{figure}

As the most representative algorithm in the field of one-stage object detection\cite{yolo1,yolo2,yolo3,yolo4,syolo4,yolo5,yolopp,yolopp2,yolof,yolox,yolo6,yolo7,yolo8,yolo9,yolo10,yolo11,focal,efficientdet,asff,ssd}, YOLO (You Only Look Once)\cite{yolo1} has demonstrated a remarkable evolutionary trajectory in its development history. Since the introduction of the YOLOv1 architecture by Joseph Redmon et al\cite{yolo1}. in 2016, the series has achieved revolutionary breakthroughs in architectural design and performance optimization up to the release of YOLO11\cite{yolo11} in 2024. From the perspective of technological evolution, the YOLO series'\cite{yolo1,yolo2,yolo3,yolo4,syolo4,yolo5,yolopp,yolopp2,yolof,yolox,yolo6,yolo7,yolo8,yolo9,yolo10,yolo11} development has been characterized by three key technical dimensions: foundational innovations in network architecture optimization and image enhancement techniques, transformative advancements in feature extraction mechanisms and loss function design, and groundbreaking paradigm shifts enabled by Transformer architecture\cite{vit} integration. Particularly noteworthy breakthroughs have been made in attention mechanism\cite{senet,cbam,attention} optimization, lightweight model\cite{efficientdet,yolo11} design, and multi-scale feature fusion\cite{csp,fpn,pan,yolocs}.

However, it is observed that since ViT\cite{vit} introduced the Transformer architecture to vision models, the core self-attention mechanism suffers from computational inefficiency. Furthermore, its channel-wise multi-head operation fails to achieve global feature attention. Additionally, the decoupled head\cite{yolox} designs in current YOLOv8\cite{yolo8} and YOLO11 architectures increase parameters and GFLOPs, consequently reducing overall inference speed. To address these limitations, we propose a novel self-attention-based bottleneck structure—the Instance-Specific Bottleneck with Full-Channel Global Self-Attention—which resolves the non-global attention problem induced by multi-head partitioning from an image-patch perspective. To mitigate the high parameter and computational costs of decoupled heads, we further introduce a lightweight alternative: the Instance-Specific Asymmetric Decoupled Head.

\begin{figure}[H]
  \centering
  \includegraphics[scale=0.45]{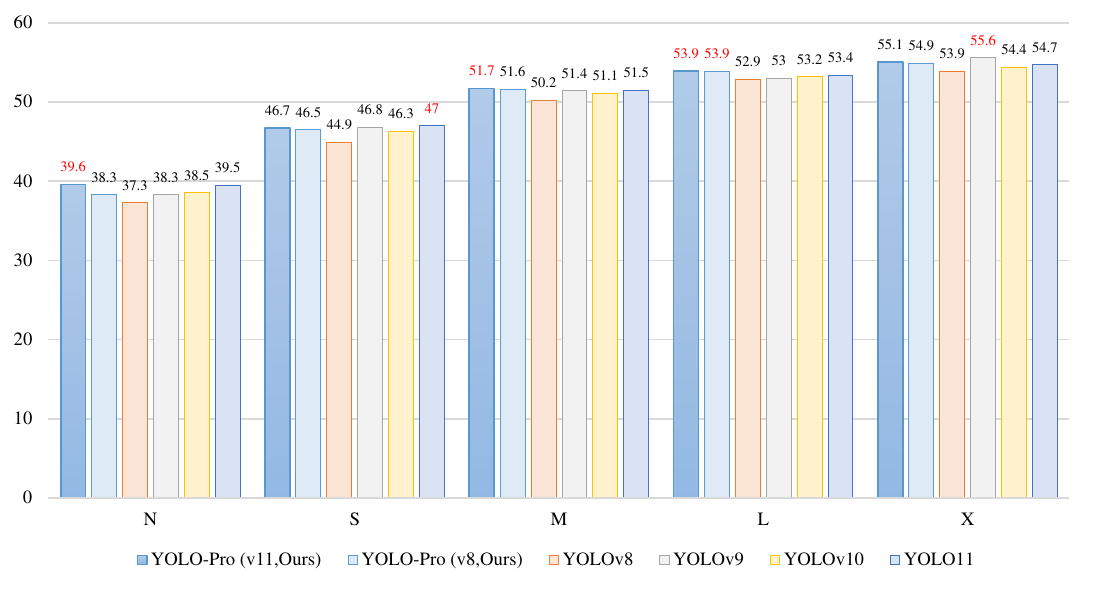}
  \caption{Comparison of the proposed YOLO-PRO with YOLOv8-YOLO11. Red numbers denote the highest AP in their corresponding categories.}
  \label{fig:ap}
\end{figure}

\textbf{Contribution}, the proposed YOLO-PRO model, built upon the YOLOv8 baseline architecture, demonstrates consistent AP improvements of 1.0\%, 1.6\%, 1.4\%, 1.0\%, and 1.1\% across N/S/M/L/X computational scale variants respectively (Fig.\ref{fig:total}). Furthermore, it achieves additional AP gains of 0.1\%, 0.5\%, and 0.2\% over YOLO11 (Fig.\ref{fig:ap}) counterparts in the critical N/M/L/X scale groups, substantiating its superior multi-scale detection capabilities. The principal contributions of this research are threefold:


1. We propose a novel bottleneck -- Instance-Specific Bottleneck with Full-Channel Global Self-Attention (ISB). This architecture implements instance normalization (Conv-IN-Activation) on convolutional layers to explicitly differentiate and amplify channel-specific discriminative features across instances. By strategically unfolding image patches into channel dimensions through spatial reorganization 
, we construct position-sensitive feature representations that intrinsically correlate full-channel information with spatial contexts. 
The proposed method establishes explicit correlations between channel features and spatial positions while maintaining linear complexity growth with respect to input size.

2. We propose a novel decoupled head -- the Instance-Specific Asymmetric Decoupled Head (ISADH).The proposed network architecture employs an asymmetric decoupled head\cite{yolocs} design, which configures convolutional kernel sizes for distinct branches based on the computational logic between the loss function and detection head\cite{yolocs}. Specifically, a 3×3 kernel is adopted for the classification branch while a 1×1 kernel is implemented for the bounding box regression branch, effectively reducing the parameters and GFLOPs of the structure. Furthermore, the architecture introduces an instance-specific feature branch that compensates for instance-wise feature variations in batch samples through feature fusion with original branches. This innovative integration mechanism achieves dual optimization objectives: significant reduction in both model parameters and GFLOPs, while maintaining competitive AP through enhanced feature representation capability.

\section{YOLO-PRO}
\subsection{Instance-Specific Bottleneck with Full-Channel Global Self-Attention (ISB)}

\begin{figure}[H]
  \centering
  \includegraphics[scale=0.38]{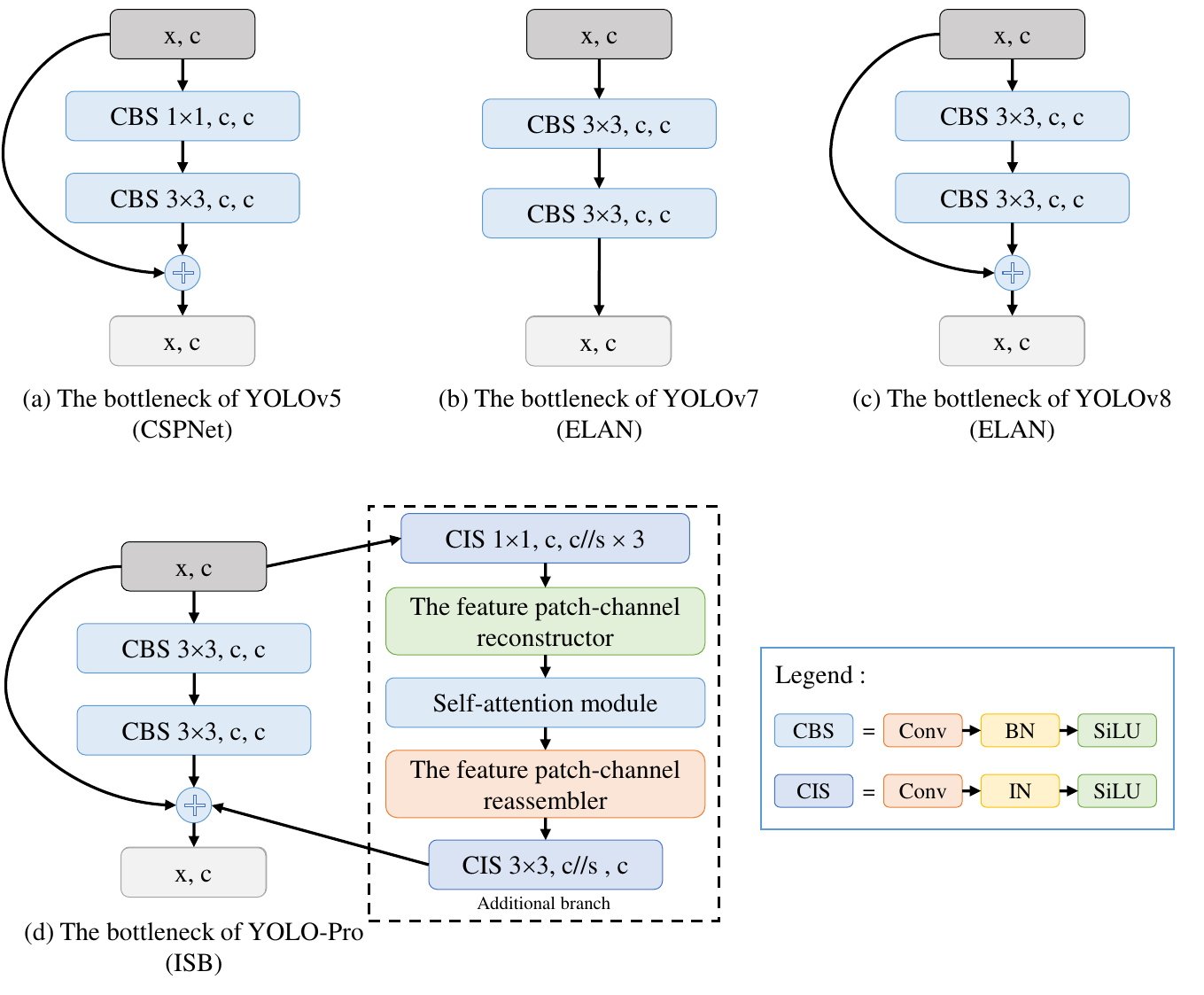}
  \caption{Bottleneck structures in backbone networks within different YOLO network architectures.}
  \label{fig:bottleneck}
\end{figure}

\begin{figure}[H]
  \centering
  \includegraphics[scale=0.38]{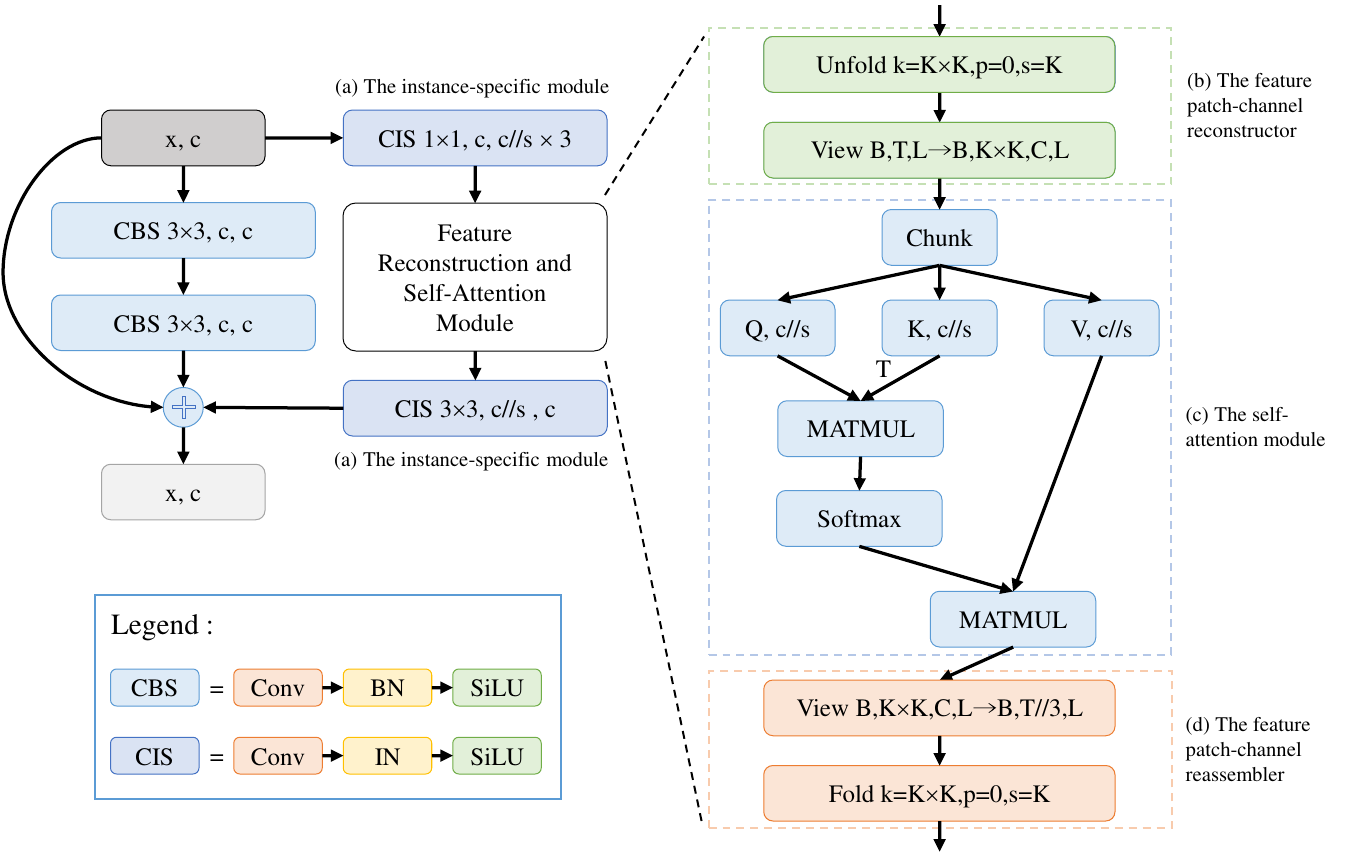}
  \caption{The structure diagram of ISB, where K (default: 4) denotes the image patch size and s (default: 8) represents the channel compression ratio.}
  \label{fig:isb}
\end{figure}

The bottleneck structure is a crucial component of both the backbone and neck networks in the YOLO series algorithms. Its primary forms include the residual block structure (Fig.\ref{fig:bottleneck} (a)) in YOLOv5, the ELAN structure (Fig.\ref{fig:bottleneck} (b)) in YOLOv7, and the improved ELAN structure (Fig.\ref{fig:bottleneck} (c)) in YOLOv8. Even YOLOv10 and YOLOv11 have adopted the bottleneck structure from YOLOv8. These structures not only enhance feature extraction capabilities through multi-level feature fusion but also improve the stability of gradient propagation, preventing gradient vanishing. Furthermore, they excel in computational efficiency and model lightweighting, significantly boosting inference speed, increasing the model's AP, and enhancing real-time inference capabilities. 
However, during our investigation of these bottleneck structures, we observed that convolutional layers in all these structures utilize BN to normalize the features of batch samples and adjust the feature distribution through trainable parameters. While this approach effectively enhances the model's generalization ability, it neglects the differences between individual sample channel features. Overemphasizing the statistical features of the batch may cause misclassification of different object categories with similar or ambiguous features (e.g., airplane and bird in the distance). This can lead to a reduction in AP during inference with new samples and cause gradient conflicts when optimizing the model’s weights during training.
Furthermore, these bottleneck structures are primarily inherited or minimally adapted from image classification networks\cite{resnet}. It should be emphasized that image classification tasks focus on identifying global semantic information of entire images, where object localization and quantity estimation are irrelevant. In contrast, object detection requires precise recognition of multiple instances with accurate spatial localization and categorical identification. Consequently, the direct transplantation or minor adaptation of classification-oriented bottleneck\cite{resnet} structures exhibits suboptimal alignment with detection objectives. The critical limitation lies in their exclusive focus on feature extraction while lacking explicit spatial positional modeling capabilities, thereby failing to establish pixel-wise global contextual correlations. This architectural deficiency directly induces positional insensitivity in detection networks.
To address these issues, we propose a novel bottleneck structure, named Instance-Specific Bottleneck with Full-Channel Global Self-Attention (ISB) (Fig.\ref{fig:isb}).

The ISB structure introduces an additional branch to the bottleneck of YOLOv8. This branch primarily aims to enhance the model’s ability to learn representations of individual instance features, as well as to improve the model’s attention to global contextual relationships while simultaneously increasing its sensitivity to spatial positional information.
Based on these two objectives, the ISB branch is designed to incorporate a structure consisting of four key components : the instance-specific module, the feature patch-channel reconstructor, the self-attention module, and the feature patch-channel reassembler.

The construction of the instance-specific module (Fig.\ref{fig:isb} (a)) is relatively straightforward. The first layer consists of a $1\times1$ convolutional layer combined with an IN layer and a SiLU\cite{silu} activation, primarily used to compress the input feature map. Assuming the number of input channels is $c\in\mathbb{R}$, this layer reduces the channel count from $c$ to $c_1$, where $c_1=\lfloor \frac{c}{s} \rfloor \times 3 $ (the factor of 3 is included for the calculation of the Q, K, and V in the self-attention module\cite{vit}). Here, $s$ is set to 8 by default, representing the compression ratio. This channel compression not only improves model efficiency but also refines the features. The second layer is located at the end of the branch and consists of a $3\times3$ convolutional layer combined with an IN layer and a SiLU activation. This layer restores the compressed channel count from $c_2=\lfloor \frac{c}{s} \rfloor$ (without multiplying by 3, as the results from the self-attention module's Q, K, and V have already been computed) to $c_3$, where $c_3=c$.
The primary objective of this module is to normalize the internal instance (channel) features of each sample independently, without relying on batch statistics. This enhances the distinct features of individual instances while avoiding potential issues related to inaccurate statistical estimates that may arise from batch normalization. Additionally, through the use of channel compression and expansion, the module effectively reduces the parameters and GFLOPs, thereby improving the inference efficiency of the model.

\begin{figure}[H]
  \centering
  \includegraphics[scale=0.85]{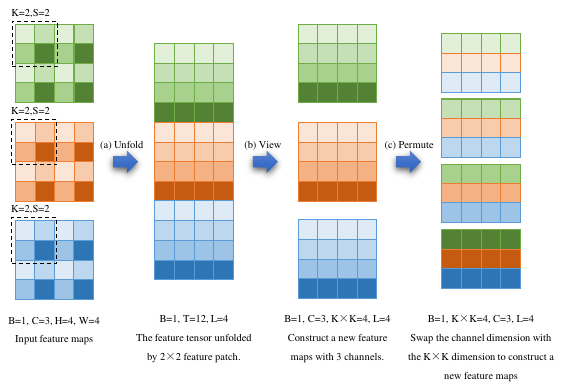}
  \caption{The feature patch-channel reconstructor's process of reconstructing the feature maps. For better understanding, the input feature map in the figure has a batch size of 1, C = 3, H = 4, and W = 4.}
  \label{fig:fpcr}
\end{figure}

The feature patch-channel reconstructor is a pivotal module designed to restructure input feature maps through a three-stage transformation. Given an input $x\in\mathbb{R}^{B \times C \times H \times W}$, it first decomposes the spatial structure by unfolding non-overlapping $K \times K$ (default is 4) patches (stride K) into a flattened tensor $x_1\in\mathbb{R}^{B \times (CK^2) \times L}$, where $L= \lfloor \frac{H -1}{K} +1 \rfloor \times \lfloor \frac{W -1}{K} +1 \rfloor $. Subsequently, the tensor is reshaped into $x_2\in\mathbb{R}^{B \times C \times K^2 \times L}$ to decouple channel and patch dimensions, followed by permuting the $K^2$ and $C$ axes to produce the final representation $x_3\in\mathbb{R}^{B \times K^2 \times C \times L}$. This hierarchical reorganization explicitly isolates spatial-local ($K^2$) and channel-wise ($C$) interactions, facilitating efficient feature reconstruction for downstream tasks.
This module aims to reconstruct features for the self-attention module, fundamentally diverging from the feature processing paradigms of Vision Transformer (ViT)\cite{vit} and Swin-Transformer\cite{swin}.

\begin{figure}[H]
  \centering
  \includegraphics[scale=0.9]{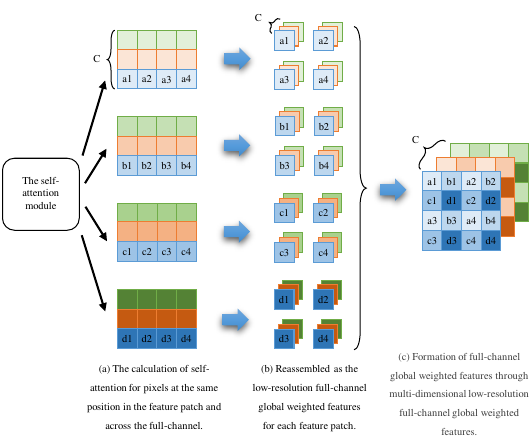}
  \caption{A schematic of the full-channel global self-attention between the feature patch pixels and the full channel.}
  \label{fig:sa}
\end{figure}

In comparison with ViT and Swin-Transformer, our approach fundamentally redesigns the construction of input self-attention feature maps by replacing the head division method in the multi-head mechanism\cite{vit} with feature patch size adaptation (Fig.\ref{fig:sa} (a)). The processed feature maps ($x_3$) are fed into the self-attention module to derive low-resolution full-channel global weighted features based on fixed positional pixels within feature patchs (Fig.\ref{fig:sa} (b)). These partial features are subsequently aggregated to form comprehensive full-channel global representations (Fig.\ref{fig:sa} (c)). To address the lack of intra-patch pixel information interaction, we implement a $3 \times 3$ convolutional layer (CIS) to facilitate cross-pixel feature sharing within each block, thereby achieving complete full-channel global attention weighting. This methodology not only effectively captures cross-channel global dependencies at the feature block level but also significantly enhances computational efficiency of the self-attention mechanism. Furthermore, considering the heightened positional sensitivity in object detection models compared to classification tasks, our proposed full-channel attention mechanism demonstrates superior capability in modeling global spatial relationships between feature blocks and their constituent pixels, ultimately improving AP in object detection.

The computational procedure of the self-attention module has been improved compared to ViT. Given input feature maps $Q, K, V\in\mathbb{R}^{B \times  K^2 \times C \times L}$ and output feature maps $F\in\mathbb{R}^{B \times  K^2 \times C \times L}$, the specific computation process is implemented as follows:

\begin{equation}
F = Softmax((\frac{Q}{\sqrt{L}})K^T)V
\end{equation}
Where the Softmax function serves as the activation function, and $L$ represents the number of feature patchs. The improvement in this equation, compared to the ViT, lies in the early division of $Q$ by the scaling factor $\sqrt{L}$. This adjustment aims to prevent the issue of excessively large values in the matrix multiplication of $Q$ and $K^T$, which can lead to NaN values during training.

The feature patch-channel reassembler operates through three key computational stages: First, the feature tensor $x_f\in\mathbb{R}^{B \times K^2 \times C \times L}$ processed by the self-attention module undergoes dimension permutation, where the $K^2$ and C dimensions are transposed to produce intermediate feature $x_{f1}\in\mathbb{R}^{B \times C \times K^2 \times L}$. Subsequently, a tensor reshaping operation is applied to transform $x_{f1}$ into a 2D feature matrix $x_{f2}\in\mathbb{R}^{B \times (CK^2) \times L}$. Finally, the fold operation reassembles $x_{f2}$ into feature maps $x_{f3}\in\mathbb{R}^{B \times C \times H \times W}$, which are then fed into convolutional layers for subsequent processing.


In summary, the ISB module is fundamentally designed to enhance instance feature discriminability by fusing instance-level features into batch-level feature representations. Meanwhile, departing from the conventional multi-head attention paradigm with subspace partitioning, ISB innovatively reconstructs feature maps into feature patchs and establishes global attention correlations across full-channel dimensions and spatially fixed positions within each patch. A convolutional layer is subsequently employed to enable cross-patch feature interaction, thereby constructing an efficient full-channel global self-attention mechanism. This structure not only strengthens contextual dependency modeling and spatial position-aware representation, but also achieves significant improvements in AP.

\subsection{Instance-Specific Asymmetric Decoupled Head (ISADH)}

The optimization of the decoupled head has been a central focus, as its need to decouple classification and regression tasks into separate network branches inherently results in a substantial increase in both model parameters and computational cost (GFLOPs), thus degrading the model's overall inference efficiency. Furthermore, since the decoupled head directly generates final detection outputs, its structural design critically impacts the  AP. To address these dual objectives, we optimize the decoupled head through architectural refinement and computational efficiency enhancements, proposing two key design innovations.

\begin{figure}[H]
  \centering
  \includegraphics[scale=0.42]{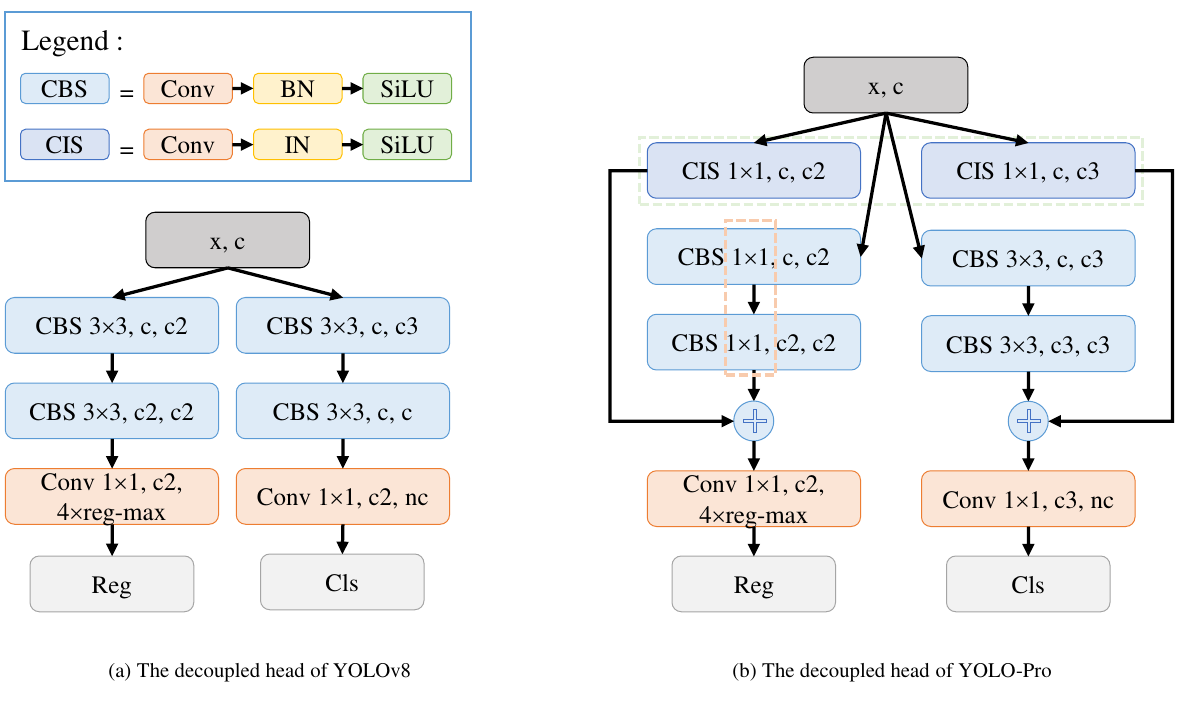}
  \caption{Comparison of Decoupled Head Structures between YOLO-PRO and YOLOv8, where C2 and C3 denote channel dimensions (consistent with YOLOv8), reg-max indicates the maximum number of regression parameters for bounding box prediction, and nc represents the number of object classes.}
  \label{fig:isadh}
\end{figure}

The first key design (The region enclosed by the orange dashed box in Fig.\ref{fig:isadh} (b)) involves reducing the kernel size of the bounding box prediction branch in the YOLOv8 decoupled head (Fig.\ref{fig:isadh} (a)) from $3 \times 3$ to $1 \times 1$. This modification significantly reduces the parameters and GFLOPs of the decoupled head while preserving AP performance. Our design rationale draws from YOLOCS’s analysis of the logical relationship between detection heads and loss functions\cite{yolocs}. In object detection, the confidence branch (merged with the classification branch in YOLOv8) primarily handles global grid predictions, making bounding box regression subordinate to classification in task priority. Consequently, strategically minimizing parameters allocated to regression tasks not only maintains detection accuracy but also enhances decoupled head efficiency.


The second key design (The region enclosed by the green dashed box in Fig.\ref{fig:isadh} (b)) involves introducing parallel instance-specific branches to both classification and regression task branches. Specifically, we construct a dedicated feature processing path comprising a $1 \times 1$ convolutional layer, instance normalization layer, and activation function within each task branch. This structure aims to extract instance-specific channel-normalized features from individual samples and adaptively fuse them with batch-level statistical features, thereby enhancing the decoupled detection head's capacity to characterize sample-unique discriminative features. The proposed dual-stream feature fusion mechanism not only incorporates global statistical information from batch normalization but also preserves instance-specific features. Through synergistic optimization of these complementary features, this approach achieves significant improvement in AP for the detection head.

﻿
In summary, these two key designs collectively constitute our proposed Instance-Specific Asymmetric Decoupled Head (ISADH) module. The structure aims to achieve dual objectives: 1) substantially improving inference efficiency by optimizing parameters and GFLOPs, and 2) significantly enhancing detection accuracy through synergistic integration of instance-specific features with batch-statistical features, as evidenced by the notable improvement in AP.

\section{Experiments}
Our experimental framework is established based on YOLOv8 and YOLO11 as the baselines, with all configurations strictly aligned with the original implementation to ensure comparability. The controlled experimental settings include four key aspects: 1) Data augmentation pipeline: Maintaining identical preprocessing strategies (Mosaic augmentation, MixUp blending, and Copy-Paste synthesis); 2) Network architecture: Preserving the structural integrity of the Backbone, Neck, and Detection Head (only replacing target modules)\cite{yolo4}; 3) Post-processing: Consistently applying Non-Maximum Suppression (NMS); 4) Hyperparameters: Fully inheriting the original training configurations. The proposed ISB and ISADH modules are directly integrated into the YOLOv8 architecture for end-to-end training and validation. For comprehensive comparison with state-of-the-art (SOTA) methods, we conduct experiments on the MS-COCO 2017 benchmark dataset\cite{mscoco} following standard splits: train2017\cite{mscoco} for model training, val2017\cite{mscoco} for performance validation, and test2017\cite{mscoco} for final evaluation. All models are trained for 500 epochs using a computational cluster equipped with 8 NVIDIA RTX3090 GPUs.

\subsection{Visualization Analysis}

The visualization analysis employed Eigen-CAM\cite{eigen} to generate attention heatmaps for representative samples selected from the val2017 dataset. The selected samples were divided into five categories: single-class single-object, single-class multi-object, single-class overlapping objects, multi-class overlapping objects, and multi-class multi-object. Attention heatmaps were generated for these samples using the baseline YOLOv8 model, the state-of-the-art YOLO11 model, and our proposed YOLO-PRO model, followed by comparative analysis.

\begin{figure}[H]
  \centering
  \includegraphics[scale=0.42]{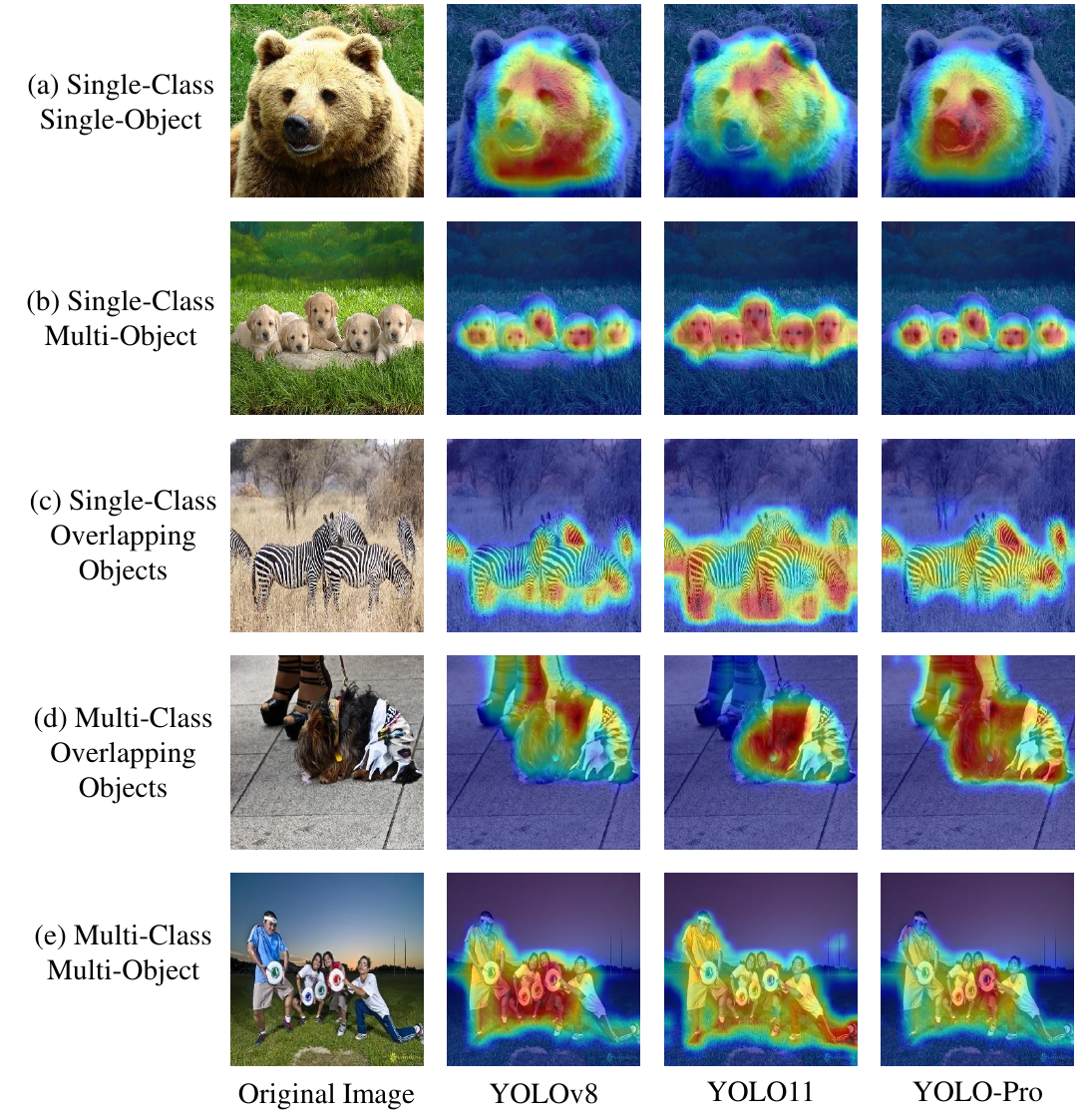}
  \caption{Attention heatmaps generated by Eigen-CAM\cite{eigen} were employed to conduct a visual analysis of diverse object categories.}
  \label{fig:va}
\end{figure}


In the single-class single-object category group (Fig.\ref{fig:va} (a)), our proposed YOLO-PRO model demonstrated more precise coverage of object feature regions in attention heatmaps compared to other models, with minimal background noise interference. In the single-class multi-object category (Fig.\ref{fig:va} (b)), the YOLOv8 model exhibited uneven attention heatmap distributions across objects, showing a bias toward centrally located objects. The YOLO11 model generated broader attention heatmap regions for each object but suffered from excessive background noise and inconsistent attention allocation to peripheral objects. In contrast, our YOLO-PRO model achieved uniform attention distribution across all objects while maintaining minimal background interference.

For the single-class overlapping objects group (Fig.\ref{fig:va} (c)), both YOLOv8 and YOLO11 displayed varying degrees of attention bias and failed to distinguish different instances within overlapping regions. The YOLO-PRO model, however, produced evenly distributed attention heatmaps that clearly differentiated overlapping instances. In the multi-class overlapping objects group (Fig.\ref{fig:va} (d)), the YOLO11 model exhibited severe attention bias toward specific classes (e.g., prioritizing "dog" detection), while YOLOv8 generated uneven heatmaps for coexisting classes. The YOLO-PRO model outperformed both, achieving balanced attention distribution without class-specific bias.

In the multi-class multi-object group (Fig.\ref{fig:va} (e)), the YOLO11 model continued to show significant attention bias. The YOLOv8 model suffered from feature confusion between adjacent objects (e.g., the leftmost and second-left human instances) and prioritized high-frequency categories (e.g., the "person" class in MS-COCO), resulting in suboptimal global attention allocation. The YOLO-PRO model avoided over-prioritizing dominant categories and mitigated suppression of small objects by large ones. While all three models exhibited uneven global attention distribution, YOLO-PRO demonstrated superior performance in suppressing excessive large-object dominance and class prioritization. The visualization analysis conclusively demonstrates the outstanding performance of YOLO-PRO.

\subsection{Ablation Experiment}


To validate the effectiveness of the proposed modules, we conducted ablation experiments on the MS-COCO dataset. Based on the baseline YOLOv8 architecture, we successively incorporated the innovative ISB and ISADH modules individually and in combination, comparing the variations in AP metrics while evaluating changes in parameters (Params) and computational cost (GFLOPs). The experimental results shown in Tab.\ref{tab:ablation} demonstrate the AP improvements over the baseline model, with green highlights indicating the enhancement magnitude.

\begin{table}[H]
\centering
	\caption{Ablation Experiment on MS-COCO val2017. The experimental data were obtained from experiments conducted using YOLOv8 as the baseline model, incorporating the proposed innovative module.}
	\label{tab:ablation}
\adjustbox{scale=0.85}{
	\begin{tabular}{ccclcc}
	\toprule
    Baseline & ISB & ISADH & AP(\%) & Params(M) & GFLOPs  \\
    \midrule
    \checkmark  & $\times$ & $\times$ & 52.9 & 43.7 & 165.7  \\
    \checkmark  & \checkmark & $\times$ & 53.8 (\textcolor{green}{+0.9})  & 45.7 & 174.0 \\
    \checkmark  & $\times$ & \checkmark & 53.1 (\textcolor{green}{+0.2})   & 43.3 & 164.2  \\
    \checkmark  & \checkmark & \checkmark & 53.9 (\textcolor{green}{+1.0}) & 45.3 & 172.4  \\
    \bottomrule
\end{tabular}
}
\end{table}


Upon integrating the ISB module into the baseline architecture, we observed a 0.9\% enhancement in the key detection metric AP, accompanied by a 2M increase in parameters and an 8.3 GFLOPs rise in computational cost. The achieved performance gain with minimal resource overhead confirms the structural efficacy of this module. Remarkably, the incorporation of the ISADH module into the baseline model not only elevated AP by 0.2\% but also reduced both parameters and computational cost by 0.4M and 1.5 GFLOPs respectively, demonstrating its optimization capabilities. The combined integration of ISB and ISADH modules ultimately yielded a 1.0\% AP improvement with 1.6M additional parameters and 6.7 GFLOPs computational increment, substantiating that the synergistic operation of dual modules achieves optimal performance enhancement within acceptable resource constraints.

\subsection{Comparative Experiment}


This study conducts a multi-scale model comparison experiment by simultaneously adjusting network depth factors and width factors while constraining the maximum number of channels, thereby constructing five computational-scale model variants (N/S/M/L/X) as detailed in Tab.\ref{tab:scale}. To ensure experimental comparability, the depth scaling ratios, width scaling ratios, and maximum channel configurations of all scale models strictly align with the YOLOv8 baseline model specifications (parameter configurations see Tab.\ref{tab:scale}). All models were trained and evaluated on the MS-COCO dataset, utilizing AP as the primary evaluation metric while comparatively analyzing key indicators including parameters (Params), GFLOPs, and inference latency. Latency measurements were conducted under standardized hardware conditions: using an NVIDIA T4 GPU with TensorRT acceleration framework and half-precision floating-point (FP16) mode. To mitigate measurement deviations caused by hardware environmental fluctuations, the final latency data was determined based on the minimum inference time per single image.

\begin{table}[H]
\centering
	\caption{Model Compound Scaling Coefficients: Depth Factor, Width Factor, and Maximum Channels Configuration.}
	\label{tab:scale}

\adjustbox{scale=0.85}{
	\begin{tabular}{cccc}
	\toprule
    Models & Depth(\%) & Width(\%) & Max Channels  \\
    \midrule
    N  & 0.33 & 0.25 & 1024  \\
    S  & 0.33 & 0.50 & 1024 \\
    M  & 0.67 & 0.75 & 768  \\
    L  & 1.00 & 1.00 & 512  \\
    X  & 1.00 & 1.25 & 512  \\
    \bottomrule
\end{tabular}
}
\end{table}


As shown in Tab.\ref{tab:comparison8}, the comparative analysis between the proposed YOLO-PRO and YOLOv8 baseline models across multiple computational scales demonstrates that YOLO-PRO variants achieve a 1.0\%-1.6\% improvement in AP over their YOLOv8 counterparts, while maintaining comparable inference latency, parameters (Params), and GFLOPs. All YOLO-PRO variants consistently outperform the baseline model in AP metrics. Remarkably, the L-scale variant of YOLO-PRO attains comparable AP to the X-scale YOLOv8 model while demonstrating superior computational efficiency. These results confirm that YOLO-PRO achieves comprehensive performance breakthroughs with preserved resource efficiency, establishing leading performance in the field.

\begin{table}[H]
\centering
	\caption{Comparison of YOLO-PRO and YOLOv8 in terms of AP on MS-COCO val2017. The experimental data were obtained from experiments conducted using YOLOv8 as the baseline model, incorporating the proposed innovative module.}
	\label{tab:comparison8}

\adjustbox{scale=0.80}{
	\begin{tabular}{llccc}
	\toprule
    Models & AP(\%) & Params(M) & GFLOPs & Latency(ms/img)\\
    \midrule
    v8-N & 37.3 & 3.2 & 8.7 & 1.47 \\
    PRO-N & 38.3 (\textcolor{green}{+1.0}) & 3.5 & 9.9 & 1.53 \\
    \hline
    v8-S & 44.9 & 11.2 & 28.6 & 2.66 \\
    PRO-S & 46.5 (\textcolor{green}{+1.6}) & 12.6 & 32.6 & 2.82 \\
    \hline
    v8-M & 50.2 & 25.9 & 78.9 & 5.86\\
    PRO-M & 51.6 (\textcolor{green}{+1.4}) & 28.2 &  87.0 & 6.59 \\
    \hline
    v8-L & 52.9 & 43.7 & 165.2 & 9.06 \\
    PRO-L & 53.9 (\textcolor{green}{+1.0}) & 45.3 & 172.4 & 10.26 \\
    \hline
    v8-X & 53.9 & 68.2 & 257.8 & 14.37 \\
    PRO-X & 54.9 (\textcolor{green}{+1.0}) & 70.8 & 268.9 & 17.07 \\
    \bottomrule
\end{tabular}
}
\end{table}


To further validate the model's advancement, this study conducts a multidimensional performance comparison between YOLO-PRO and the state-of-the-art (SOTA) YOLO11(baseline) model (Tab.\ref{tab:comparison11}, where red highlighting indicates AP disadvantages and green highlighting denotes advantages). Experimental results show that YOLO-PRO variants (N/M/L/X) outperform their YOLO11 counterparts by 0.1\%-0.5\% in AP. Crucially, YOLO-PRO maintains persistent superiority in the core detection accuracy metric, which substantiates its SOTA effectiveness in object detection tasks.

\begin{table}[H]
\centering
	\caption{Comparison of YOLO-PRO and YOLO11 in terms of AP on MS-COCO val2017. The experimental data were obtained from experiments conducted using YOLO11 as the baseline model, incorporating the proposed innovative module.}
	\label{tab:comparison11}

\adjustbox{scale=0.80}{
	\begin{tabular}{llccc}
	\toprule
    Models & AP(\%) & Params(M) & GFLOPs & Latency(ms/img)\\
    \midrule
    11-N & 39.5 & 2.6 & 6.5 & 1.50 \\
    PRO-N & 39.6 (\textcolor{green}{+0.1}) & 2.9 & 6.4 & 1.59 \\
    \hline
    11-S & 46.7 & 9.4 & 21.5 & 2.50 \\
    PRO-S & 46.7 (\textcolor{red}{0}) & 10.9 & 22.5 & 2.68 \\
    \hline
    11-M & 51.5 & 20.1 & 68.0 & 4.70 \\
    PRO-M & 51.7 (\textcolor{green}{+0.2}) & 22.9 &  70.9 & 5.19 \\
    \hline
    11-L & 53.4 & 25.3 & 86.9 & 6.20 \\
    PRO-L & 53.9 (\textcolor{green}{+0.5}) & 28.2 & 90.1 & 8.33 \\
    \hline
    11-X & 54.7 & 56.9 & 194.9 & 11.30 \\
    PRO-X & 55.1 (\textcolor{green}{+0.4}) & 63.2 & 200.0 & 13.41 \\
    \bottomrule
\end{tabular}
}
\end{table}

\subsection{Comparison with SOTA}


As presented in Tab.\ref{tab:sota}, this study conducts a systematic comparative analysis between the proposed YOLO-PRO and contemporary SOTA object detectors within a unified benchmarking framework. The evaluation encompasses five critical dimensions: primary detection accuracy metric (AP), input resolution, parameters (Params), GFLOPs, and inference latency. Notably, boldface numerical values in the table denote detectors achieving optimal AP performance within their respective computational scale groups (N/S/M/L/X).

\begin{table}[H]
\centering
	\caption{Comparative Evaluation of AP and Inference Latency Across SOTA Object Detectors on MS-COCO val2017. }
	\label{tab:sota}

\adjustbox{scale=0.7}{
	\begin{tabular}{lccccc}
	\toprule
    Models &  Size & Params(M) & GFLOPs & Latency(ms/img) & AP(\%)  \\
    \midrule
     YOLOv8-N & 640 & 3.2 & 8.7 & 1.47 & 37.3 \\
     YOLOv9-T & 640 & 2.0 & 7.7 & 2.30 & 38.3 \\
     YOLOv10-N & 640 & - & 6.7 & 1.56 & 38.5 \\
     YOLO11-N & 640 & 2.6 & 6.5 & 1.50 & 39.5 \\
     YOLO-PRO-N(v8) & 640 & 3.5 & 9.9 & 1.53 & 38.3 \\
     YOLO-PRO-N(v11) & 640 & 2.9 & 6.4 & 1.59 & \textbf{39.6} \\
     \hline
     YOLOv8-S & 640 & 11.2 & 28.6 & 2.66 & 44.9 \\
     YOLOv9-S & 640 & 7.2 & 26.7 & 3.54 & 46.8 \\
     YOLOv10-S & 640 & - & 21.6 & 2.66 & 46.3 \\
     YOLO11-S & 640 & 9.4 & 21.5 & 2.50 & \textbf{47.0} \\
     YOLO-PRO-S(v8) & 640 & 12.6 & 32.6 & 2.82 & 46.5 \\
     YOLO-PRO-S(v11) & 640 & 10.9 & 22.5 & 2.68 & 46.7 \\
     \hline
     YOLOv8-M & 640 & 25.9 & 78.9 & 5.86 & 50.2 \\
     YOLOv9-M & 640 & 20.1 & 76.8 & 6.43 & 51.4 \\
     YOLOv10-M & 640 & - & 59.1 & 5.48 & 51.1 \\
     YOLO11-M & 640 & 20.1 & 68.0 & 4.70 & 51.5 \\
     YOLO-PRO-M(v8) & 640 & 28.2 & 87.0 & 6.59 & 51.6 \\
     YOLO-PRO-M(v11) & 640 & 22.9 &  70.9 & 5.19 & \textbf{51.7} \\
     \hline
     YOLOv8-L & 640 & 43.7 & 165.2 & 9.06 & 52.9 \\
     YOLOv9-C & 640 & 25.5 & 102.8 & 7.16 & 53.0 \\
     YOLOv10-L & 640 & - & 120.3 & 8.33 & 53.2 \\
     YOLO11-L & 640 & 25.3 & 86.9 & 6.20 & 53.4 \\
     YOLO-PRO-L(v8) & 640 & 45.3 & 172.4 & 10.26 & \textbf{53.9} \\
     YOLO-PRO-L(v11) & 640 & 28.2 & 90.1 & 8.33 & \textbf{53.9} \\
     \hline
     YOLOv8-X & 640 & 68.2 & 257.8 & 14.37 & 53.9 \\
     YOLOv9-E & 640 & 58.1 & 192.5 & 16.77 & \textbf{55.6} \\
     YOLOv10-X & 640 & - & 160.4 & 12.20 & 54.4 \\
     YOLO11-X & 640 & 56.9 & 194.9 & 11.30 & 54.7 \\
     YOLO-PRO-X(v8) & 640 & 70.8 & 268.9 & 17.07 & 54.9 \\
     YOLO-PRO-X(v11) & 640 & 63.2 & 200.0 & 13.41 & 55.1 \\
     \bottomrule
\end{tabular}
}
\end{table}


As demonstrated by the experimental data in Tab.\ref{tab:sota}, the proposed YOLO-PRO achieves optimal AP in medium (M) and large (L) computational scale groups while maintaining comparable parameters (Params), GFLOPs, and inference latency relative to peer models. This systematic comparison of core metrics confirms the architectural advancement of YOLO-PRO in object detection tasks, substantiating its SOTA status.

\section{Conclusion}


This study systematically investigates the inherent limitations of conventional bottleneck structures and decoupled heads, proposing two innovative solutions: the Instance-Specific Bottleneck with Full-Channel Global Self-Attention (ISB) module, and the Instance-Specific Asymmetric Decoupled Head (ISADH) module. The ISB module synergistically integrates batch-statistical features with instance-specific features through feature map reconstruction, establishing an efficient full-channel global self-attention mechanism. In parallel, the ISADH module pioneers an asymmetric decoupled head architecture for hierarchical fusion of multi-dimensional features by synergistically integrating batch-statistical features with instance-specific representations. Extensive experimentation confirms that the coordinated deployment of both modules effectively overcomes existing performance bottlenecks while maintaining computational efficiency, ultimately establishing new SOTA benchmarks in object detection.

\bibliography{yolopro.bib}

\begin{thebibliography}{41}
\providecommand{\natexlab}[1]{#1}

\bibitem[{Antipov et~al.(2017)Antipov, Baccouche, Berrani, and Dugelay}]{pr2}
Antipov, G.; Baccouche, M.; Berrani, S.-A.; and Dugelay, J.-L. 2017.
\newblock Effective training of convolutional neural networks for face-based
  gender and age prediction.
\newblock \emph{Pattern Recognition}, 72: 15--26.

\bibitem[{Ao~Wang(2024)}]{yolo10}
Ao~Wang, L. L. e.~a., Hui~Chen. 2024.
\newblock YOLOv10: Real-Time End-to-End Object Detection.
\newblock \emph{arXiv preprint arXiv:2405.14458}.

\bibitem[{Bochkovskiy, Wang, and Liao(2020)}]{yolo4}
Bochkovskiy, A.; Wang, C.-Y.; and Liao, H.-Y.~M. 2020.
\newblock Yolov4: Optimal speed and accuracy of object detection.
\newblock \emph{arXiv preprint arXiv:2004.10934}.

\bibitem[{Chen et~al.(2021)Chen, Wang, Yang, Zhang, Cheng, and Sun}]{yolof}
Chen, Q.; Wang, Y.; Yang, T.; Zhang, X.; Cheng, J.; and Sun, J. 2021.
\newblock You only look one-level feature.
\newblock In \emph{Proceedings of the IEEE/CVF conference on computer vision
  and pattern recognition}, 13039--13048.

\bibitem[{Dosovitskiy et~al.(2020)Dosovitskiy, Beyer, Kolesnikov, Weissenborn,
  Zhai, Unterthiner, Dehghani, Minderer, Heigold, Gelly et~al.}]{vit}
Dosovitskiy, A.; Beyer, L.; Kolesnikov, A.; Weissenborn, D.; Zhai, X.;
  Unterthiner, T.; Dehghani, M.; Minderer, M.; Heigold, G.; Gelly, S.; et~al.
  2020.
\newblock An image is worth 16x16 words: Transformers for image recognition at
  scale.
\newblock \emph{arXiv preprint arXiv:2010.11929}.

\bibitem[{Ge et~al.(2021)Ge, Liu, Wang, Li, and Sun}]{yolox}
Ge, Z.; Liu, S.; Wang, F.; Li, Z.; and Sun, J. 2021.
\newblock Yolox: Exceeding yolo series in 2021.
\newblock \emph{arXiv preprint arXiv:2107.08430}.

\bibitem[{Ghiasi et~al.(2021)Ghiasi, Cui, Srinivas, Qian, Lin, Cubuk, Le, and
  Zoph}]{copypaste}
Ghiasi, G.; Cui, Y.; Srinivas, A.; Qian, R.; Lin, T.-Y.; Cubuk, E.~D.; Le,
  Q.~V.; and Zoph, B. 2021.
\newblock Simple copy-paste is a strong data augmentation method for instance
  segmentation.
\newblock In \emph{Proceedings of the IEEE/CVF conference on computer vision
  and pattern recognition}, 2918--2928.

\bibitem[{Guan et~al.(2024)Guan, Yap, Bozoki, and Liu}]{pr3}
Guan, H.; Yap, P.-T.; Bozoki, A.; and Liu, M. 2024.
\newblock Federated learning for medical image analysis: A survey.
\newblock \emph{Pattern Recognition}, 110424.

\bibitem[{He et~al.(2015)He, Zhang, Ren, and Sun}]{spp}
He, K.; Zhang, X.; Ren, S.; and Sun, J. 2015.
\newblock Spatial pyramid pooling in deep convolutional networks for visual
  recognition.
\newblock \emph{IEEE transactions on pattern analysis and machine
  intelligence}, 37(9): 1904--1916.

\bibitem[{He et~al.(2016)He, Zhang, Ren, and Sun}]{resnet}
He, K.; Zhang, X.; Ren, S.; and Sun, J. 2016.
\newblock Deep residual learning for image recognition.
\newblock In \emph{Proceedings of the IEEE conference on computer vision and
  pattern recognition}, 770--778.

\bibitem[{Hu, Shen, and Sun(2018)}]{senet}
Hu, J.; Shen, L.; and Sun, G. 2018.
\newblock Squeeze-and-excitation networks.
\newblock In \emph{Proceedings of the IEEE conference on computer vision and
  pattern recognition}, 7132--7141.

\bibitem[{Huang et~al.(2025)Huang, Li, Tan, Shen, Yu, and Fu}]{yolocs}
Huang, L.; Li, W.; Tan, Y.; Shen, L.; Yu, J.; and Fu, H. 2025.
\newblock YOLOCS: Object detection based on dense channel compression for
  feature spatial solidification.
\newblock \emph{Knowledge-Based Systems}, 113024.

\bibitem[{Huang et~al.(2021)Huang, Wang, Lv, Bai, Long, Deng, Dang, Han, Liu,
  Hu et~al.}]{yolopp2}
Huang, X.; Wang, X.; Lv, W.; Bai, X.; Long, X.; Deng, K.; Dang, Q.; Han, S.;
  Liu, Q.; Hu, X.; et~al. 2021.
\newblock PP-YOLOv2: A practical object detector.
\newblock \emph{arXiv preprint arXiv:2104.10419}.

\bibitem[{Jocher(2020)}]{yolo5}
Jocher, G. 2020.
\newblock Ultralytics YOLOv5.

\bibitem[{Jocher, Chaurasia, and Qiu(2023)}]{yolo8}
Jocher, G.; Chaurasia, A.; and Qiu, J. 2023.
\newblock Ultralytics YOLOv8.

\bibitem[{Jocher and Qiu(2024)}]{yolo11}
Jocher, G.; and Qiu, J. 2024.
\newblock Ultralytics YOLO11.

\bibitem[{Kim et~al.(2018)Kim, Kook, Sun, Kang, and Ko}]{fpn}
Kim, S.-W.; Kook, H.-K.; Sun, J.-Y.; Kang, M.-C.; and Ko, S.-J. 2018.
\newblock Parallel feature pyramid network for object detection.
\newblock In \emph{Proceedings of the European conference on computer vision
  (ECCV)}, 234--250.

\bibitem[{Li et~al.(2022)Li, Li, Jiang, Weng, Geng, Li, Ke, Li, Cheng, Nie
  et~al.}]{yolo6}
Li, C.; Li, L.; Jiang, H.; Weng, K.; Geng, Y.; Li, L.; Ke, Z.; Li, Q.; Cheng,
  M.; Nie, W.; et~al. 2022.
\newblock YOLOv6: A single-stage object detection framework for industrial
  applications.
\newblock \emph{arXiv preprint arXiv:2209.02976}.

\bibitem[{Lin et~al.(2017)Lin, Goyal, Girshick, He, and Doll{\'a}r}]{focal}
Lin, T.-Y.; Goyal, P.; Girshick, R.; He, K.; and Doll{\'a}r, P. 2017.
\newblock Focal loss for dense object detection.
\newblock In \emph{Proceedings of the IEEE international conference on computer
  vision}, 2980--2988.

\bibitem[{Lin et~al.(2014)Lin, Maire, Belongie, Hays, Perona, Ramanan,
  Doll{\'a}r, and Zitnick}]{mscoco}
Lin, T.-Y.; Maire, M.; Belongie, S.; Hays, J.; Perona, P.; Ramanan, D.;
  Doll{\'a}r, P.; and Zitnick, C.~L. 2014.
\newblock Microsoft coco: Common objects in context.
\newblock In \emph{Computer Vision--ECCV 2014: 13th European Conference,
  Zurich, Switzerland, September 6-12, 2014, Proceedings, Part V 13}, 740--755.
  Springer.

\bibitem[{Liu, Huang, and Wang(2019)}]{asff}
Liu, S.; Huang, D.; and Wang, Y. 2019.
\newblock Learning spatial fusion for single-shot object detection.
\newblock \emph{arXiv preprint arXiv:1911.09516}.

\bibitem[{Liu et~al.(2018)Liu, Qi, Qin, Shi, and Jia}]{pan}
Liu, S.; Qi, L.; Qin, H.; Shi, J.; and Jia, J. 2018.
\newblock Path aggregation network for instance segmentation.
\newblock In \emph{Proceedings of the IEEE conference on computer vision and
  pattern recognition}, 8759--8768.

\bibitem[{Liu et~al.(2016)Liu, Anguelov, Erhan, Szegedy, Reed, Fu, and
  Berg}]{ssd}
Liu, W.; Anguelov, D.; Erhan, D.; Szegedy, C.; Reed, S.; Fu, C.-Y.; and Berg,
  A.~C. 2016.
\newblock Ssd: Single shot multibox detector.
\newblock In \emph{Computer Vision--ECCV 2016: 14th European Conference,
  Amsterdam, The Netherlands, October 11--14, 2016, Proceedings, Part I 14},
  21--37. Springer.

\bibitem[{Liu et~al.(2021)Liu, Lin, Cao, Hu, Wei, Zhang, Lin, and Guo}]{swin}
Liu, Z.; Lin, Y.; Cao, Y.; Hu, H.; Wei, Y.; Zhang, Z.; Lin, S.; and Guo, B.
  2021.
\newblock Swin transformer: Hierarchical vision transformer using shifted
  windows.
\newblock In \emph{Proceedings of the IEEE/CVF international conference on
  computer vision}, 10012--10022.

\bibitem[{Long et~al.(2020)Long, Deng, Wang, Zhang, Dang, Gao, Shen, Ren, Han,
  Ding et~al.}]{yolopp}
Long, X.; Deng, K.; Wang, G.; Zhang, Y.; Dang, Q.; Gao, Y.; Shen, H.; Ren, J.;
  Han, S.; Ding, E.; et~al. 2020.
\newblock PP-YOLO: An effective and efficient implementation of object
  detector.
\newblock \emph{arXiv preprint arXiv:2007.12099}.

\bibitem[{Muhammad and Yeasin(2020)}]{eigen}
Muhammad, M.~B.; and Yeasin, M. 2020.
\newblock Eigen-cam: Class activation map using principal components.
\newblock In \emph{2020 international joint conference on neural networks
  (IJCNN)}, 1--7. IEEE.

\bibitem[{Qian, Lai, and Li(2022)}]{pr1}
Qian, R.; Lai, X.; and Li, X. 2022.
\newblock 3D object detection for autonomous driving: A survey.
\newblock \emph{Pattern Recognition}, 130: 108796.

\bibitem[{Ramachandran, Zoph, and Le(2017)}]{silu}
Ramachandran, P.; Zoph, B.; and Le, Q.~V. 2017.
\newblock Searching for activation functions.
\newblock \emph{arXiv preprint arXiv:1710.05941}.

\bibitem[{Redmon et~al.(2016)Redmon, Divvala, Girshick, and Farhadi}]{yolo1}
Redmon, J.; Divvala, S.; Girshick, R.; and Farhadi, A. 2016.
\newblock You only look once: Unified, real-time object detection.
\newblock In \emph{Proceedings of the IEEE conference on computer vision and
  pattern recognition}, 779--788.

\bibitem[{Redmon and Farhadi(2017)}]{yolo2}
Redmon, J.; and Farhadi, A. 2017.
\newblock YOLO9000: better, faster, stronger.
\newblock In \emph{Proceedings of the IEEE conference on computer vision and
  pattern recognition}, 7263--7271.

\bibitem[{Redmon and Farhadi(2018)}]{yolo3}
Redmon, J.; and Farhadi, A. 2018.
\newblock Yolov3: An incremental improvement.
\newblock \emph{arXiv preprint arXiv:1804.02767}.

\bibitem[{Tan, Pang, and Le(2020)}]{efficientdet}
Tan, M.; Pang, R.; and Le, Q.~V. 2020.
\newblock Efficientdet: Scalable and efficient object detection.
\newblock In \emph{Proceedings of the IEEE/CVF conference on computer vision
  and pattern recognition}, 10781--10790.

\bibitem[{Ulyanov, Vedaldi, and Lempitsky(2016)}]{in}
Ulyanov, D.; Vedaldi, A.; and Lempitsky, V. 2016.
\newblock Instance normalization: The missing ingredient for fast stylization.
\newblock \emph{arXiv preprint arXiv:1607.08022}.

\bibitem[{Vaswani et~al.(2017)Vaswani, Shazeer, Parmar, Uszkoreit, Jones,
  Gomez, Kaiser, and Polosukhin}]{attention}
Vaswani, A.; Shazeer, N.; Parmar, N.; Uszkoreit, J.; Jones, L.; Gomez, A.~N.;
  Kaiser, {\L}.; and Polosukhin, I. 2017.
\newblock Attention is all you need.
\newblock \emph{Advances in neural information processing systems}, 30.

\bibitem[{Wang et~al.(2024)Wang, Zhu, Gao, Gan, Zhang, Gu, Qian, Chen, and
  Ma}]{pr4}
Wang, C.; Zhu, W.; Gao, B.-B.; Gan, Z.; Zhang, J.; Gu, Z.; Qian, S.; Chen, M.;
  and Ma, L. 2024.
\newblock Real-iad: A real-world multi-view dataset for benchmarking versatile
  industrial anomaly detection.
\newblock In \emph{Proceedings of the IEEE/CVF Conference on Computer Vision
  and Pattern Recognition}, 22883--22892.

\bibitem[{Wang, Bochkovskiy, and Liao(2021)}]{syolo4}
Wang, C.-Y.; Bochkovskiy, A.; and Liao, H.-Y.~M. 2021.
\newblock Scaled-yolov4: Scaling cross stage partial network.
\newblock In \emph{Proceedings of the IEEE/cvf conference on computer vision
  and pattern recognition}, 13029--13038.

\bibitem[{Wang, Bochkovskiy, and Liao(2022)}]{yolo7}
Wang, C.-Y.; Bochkovskiy, A.; and Liao, H.-Y.~M. 2022.
\newblock YOLOv7: Trainable bag-of-freebies sets new state-of-the-art for
  real-time object detectors.
\newblock \emph{arXiv preprint arXiv:2207.02696}.

\bibitem[{Wang and Liao(2024)}]{yolo9}
Wang, C.-Y.; and Liao, H.-Y.~M. 2024.
\newblock YOLOv9: Learning What You Want to Learn Using Programmable Gradient
  Information.

\bibitem[{Wang et~al.(2020)Wang, Liao, Wu, Chen, Hsieh, and Yeh}]{csp}
Wang, C.-Y.; Liao, H.-Y.~M.; Wu, Y.-H.; Chen, P.-Y.; Hsieh, J.-W.; and Yeh,
  I.-H. 2020.
\newblock CSPNet: A new backbone that can enhance learning capability of CNN.
\newblock In \emph{Proceedings of the IEEE/CVF conference on computer vision
  and pattern recognition workshops}, 390--391.

\bibitem[{Woo et~al.(2018)Woo, Park, Lee, and Kweon}]{cbam}
Woo, S.; Park, J.; Lee, J.-Y.; and Kweon, I.~S. 2018.
\newblock Cbam: Convolutional block attention module.
\newblock In \emph{Proceedings of the European conference on computer vision
  (ECCV)}, 3--19.

\bibitem[{Zhang et~al.(2017)Zhang, Cisse, Dauphin, and Lopez-Paz}]{mixup}
Zhang, H.; Cisse, M.; Dauphin, Y.~N.; and Lopez-Paz, D. 2017.
\newblock mixup: Beyond empirical risk minimization.
\newblock \emph{arXiv preprint arXiv:1710.09412}.

\end{thebibliography}


\end{document}